\pdfoutput=1
\PassOptionsToPackage{numbers,compress,sort}{natbib}
\documentclass{article}
\usepackage[dblblindworkshop,final]{neurips_2026}
\workshoptitle{GlobalSouthAI}

\usepackage[utf8]{inputenc}
\usepackage[T1]{fontenc}
\usepackage[hidelinks]{hyperref}
\usepackage{url}
\usepackage{booktabs}
\usepackage{amsmath}
\usepackage{amsfonts}
\usepackage{nicefrac}
\usepackage{microtype}
\usepackage{xcolor}
\usepackage{graphicx}
\usepackage{float}

\newcommand{\repo}{\url{https://github.com/InnocentAnyaele/respisound}}

\title{Feature Scaling Beats Architecture Choice in Offline Respiratory Screening for Low-Resource Clinics}

\author{%
  Abubakar Chilala \quad Cleophas Kadima \quad Innocent Anyaele \quad Victor Olufemi\\
  College of Engineering, Carnegie Mellon University Africa\\
  Carnegie Mellon University, 15213, Pittsburgh, PA, USA\\
  \texttt{\{achilala, kkadima, ianyaele, volufemi\}@andrew.cmu.edu}
}

\begin{document}
\maketitle

\begin{abstract}
Deep models are the usual choice for respiratory sound classification, but in clinics without GPUs or connectivity that choice carries a cost. We compare classical and deep classifiers on a five-class respiratory-sound corpus of 1,211 clips (asthma, COPD, pneumonia, Bronchial, healthy) from 201 patient groups, with a 3.86:1 class imbalance, using pre-registered patient-grouped cross-validation (25 runs per configuration). An SVM over 30 summary features (MFCC, ZCR and RMS statistics) reaches macro recall 0.78 and 0.71 recall on the rarest class. A 1.13M-parameter convolutional recurrent network over log-Mel spectrograms reaches 0.75 to 0.77, and no difference between model families is resolved. The largest single effect we measure is a preprocessing effect rather than an architectural one. Standardizing the SVM's features, a routine step for RBF kernels, moves its macro recall from 0.51 to 0.78, a 27-point swing wider than any gap between model families here. Augmentation did not help: SMOTE, waveform augmentation and SpecAugment have no measurable effect.  Latency does not separate the models (51 to 74\,ms per clip on a CPU-only laptop), but the classical models avoid a 449\,MB PyTorch runtime.  Clip-level splits, which let a patient's recordings fall on both sides, inflate macro recall by 7.5 to 9.8 points. Recording source is strongly tied to label (Cram\'er's $V=0.65$ between native sampling rate and class), and in an exploratory analysis clips recorded at a class's minority rate are mostly assigned to classes typical of that rate; we state this confound as the paper's main limitation.
\end{abstract}

\section{Introduction}

Respiratory disease places an outsized burden on African health systems. Asthma caused over 455,000 deaths worldwide in 2019, concentrated in low- and middle-income countries (LMICs)~\cite{who_asthma}; lower respiratory infections such as pneumonia killed about 2.2 million people in 2021, including about 500,000 children under five, half of them in low-SDI countries~\cite{gbd2021}; and COPD was the fourth leading cause of death worldwide in 2021, with close to 90\% of premature COPD deaths in LMICs~\cite{who_copd}. Confirming these diagnoses relies on tools such as spirometry that are often unavailable in LMICs~\cite{who_copd}, in a region with 1.55 physicians, nurses and midwives per 1,000 people, about a third of the WHO threshold of 4.45~\cite{who_workforce}.

Respiratory sounds are an attractive signal because they need only a microphone or an electronic stethoscope~\cite{rocha}. Spectrogram-based deep models are the usual choice for classifying them~\cite{basu,moummad}, and on public corpora they report near-ceiling accuracy: 98.9\% on the Asthma Detection Dataset v2 (ADD~v2) with a stratified clip-level split, but 75.7\% across institutions~\cite{tawfik}. Without a GPU the choice of model carries a cost, so it matters whether the deep model is more accurate at the data scale available. We compare classical and deep models under a pre-registered, patient-grouped protocol with 25 runs per configuration. No ranking between model families is resolved, but one preprocessing step outweighs the choice of architecture, and a larger problem appears: recording source is entangled with the label.

\textbf{Contributions.} (i)~A pre-registered, patient-grouped comparison of SVM, RF and CRNN with paired intervals: no architecture difference is resolved, while omitting feature standardization costs 27 points. (ii)~Measured evaluation effects: clip-level leakage inflates macro recall by 7.5--9.8 points, and SMOTE, waveform augmentation and SpecAugment have no measurable effect. (iii)~Evidence that recording source is confounded with label in ADD~v2, with a cheap diagnostic that applies to any corpus. (iv)~Implications for offline deployment.

\section{Related Work}

Respiratory sound classification moved from classical models on hand-crafted descriptors~\cite{rocha} to Mel-spectrogram CNNs and recurrent models~\cite{basu,moummad}. Record-wise splits overstate clinical accuracy~\cite{saeb}, and models trained on pooled sources can learn the source instead of the pathology, as with hospital identity in chest radiographs~\cite{zech} and dataset origin in COVID-19 radiography~\cite{degrave}, a form of shortcut learning~\cite{geirhos}. ADD~v2~\cite{musaed} is used in recent work~\cite{tawfik}, but a published comment questions its relationship to the ICBHI 2017 database it is named after~\cite{dolu}.
Work on point-of-care diagnostics in African settings~\cite{quinn} and on demographic skew in clinical corpora~\cite{groh} motivates our emphasis on provenance, since accuracy on a public corpus of undocumented origin need not transfer to the population a system serves.

\section{Method}

\textbf{Data.} We use the Asthma Detection Dataset v2~\cite{musaed}, an open-access Kaggle repository, giving 1,211 recordings from 201 patient groups; labels were lowercased and stripped to prevent class fragmentation, and no recording was entirely silent, so none were dropped. The distribution is skewed, with COPD at 401 clips (33.1\%), asthma 288 (23.8\%), pneumonia 285 (23.5\%), healthy 133 (11.0\%) and Bronchial 104 (8.6\%), a max:min ratio of 3.86:1. The source carries no geographic or demographic metadata (Section~6).

\textbf{Features.} Audio was resampled to 16\,kHz. For the classical models we extracted a 30-dimensional vector of means and standard deviations over 13 MFCCs, zero-crossing rate and RMS energy, computed over the whole clip. For the CRNN we computed a log-Mel spectrogram ($n_\text{fft}=2048$, hop 512, 128 Mel bins, power-to-dB, per-clip standardization) over either the first 1.5\,s ($128\times47$) or the full 6.0\,s window ($128\times188$; 91\% of clips last exactly 6.0\,s), zero-padding shorter clips.

\textbf{Models.} SVM: RBF kernel, $C=1$, balanced class weights, with or without standardization. RF: 100 trees, with or without balanced class weights. CRNN: two convolutional blocks (32 and 64 filters, $3\times3$, batch normalization, ReLU, $2\times2$ max-pooling), dropout 0.3, a single unidirectional LSTM with 128 hidden units reading the pooled feature map along time, and a linear head over the final timestep, giving 1,134,789 parameters. Training uses Adam ($10^{-3}$), batch 32, class weights from the training labels and up to 50 epochs, with early stopping (patience 10) on a patient-grouped validation fold and the best checkpoint restored. We also test SMOTE~\cite{chawla} on the minority class, one waveform-augmented copy per training clip, and SpecAugment~\cite{park} masks without time warping.

\textbf{Protocol.} The analysis plan, data manifest and folds were committed before any model was run\footnote{\raggedright\repo. The plan, manifest and folds are frozen at commit \texttt{69ab10f}; code, results and per-fold predictions are at commit \texttt{eadf374} (branch \texttt{camera-ready-revision}). Deviations from the plan are listed in Appendix~A.}. All configurations use the same patient-grouped, class-stratified 5-fold CV repeated with five seeds (25 runs). Patient groups come from filename patient numbers, which appear to restart within each class (asthma 58, COPD 51, pneumonia 46, healthy 25, Bronchial 21); twelve files are byte-identical to others, eight under a different patient number, so we merge the groups involved. Everything fitted from data (scaler, SMOTE, augmentation, class weights) is fitted within training folds, and test folds are never augmented. The primary metric is macro recall. Paired differences over the 25 runs get 95\% intervals from the corrected resampled $t$-test~\cite{nadeau}, with wording fixed in advance: \emph{outperforms} if the interval excludes zero, \emph{comparable} if it lies within $\pm3$ points, otherwise \emph{inconclusive}. Clip-level CV measures leakage. As controls, we low-pass all audio at 1.8\,kHz, removing the band 4\,kHz recordings lack, and fit a classifier on native rate alone.

\section{Results}

\begin{table}[t]
\caption{Macro recall, mean $\pm$ SD over 25 runs. \emph{Grouped}: patient-grouped CV (primary). \emph{Clip}: clip-level CV.}
\label{tab:main}
\centering\small
\begin{tabular}{llccc}
\toprule
Model & Configuration & Grouped & Clip & Bronchial (grouped) \\
\midrule
SVM & unscaled & 0.510 $\pm$ 0.053 & 0.542 & 0.48 \\
SVM & standardized & 0.779 $\pm$ 0.034 & 0.854 & 0.71 \\
SVM & standardized, first 1.5 s & 0.742 $\pm$ 0.051 & 0.800 & 0.65 \\
RF & unweighted & 0.767 $\pm$ 0.048 & 0.870 & 0.69 \\
RF & balanced weights & 0.766 $\pm$ 0.047 & 0.864 & 0.68 \\
CRNN & 1.5 s window & 0.749 $\pm$ 0.042 & 0.829 & 0.71 \\
CRNN & 6.0 s window & 0.766 $\pm$ 0.045 & n/a & 0.67 \\
\bottomrule
\end{tabular}
\end{table}

\begin{table}[t]
\caption{Paired macro-recall differences (points), 95\% corrected intervals; all 17 in Appendix~C.}
\label{tab:compare}
\centering\small
\begin{tabular}{lccl}
\toprule
Comparison & $\Delta$ & 95\% CI & Verdict \\
\midrule
Standardization (SVM) & +26.9 & [+22.1, +31.6] & outperforms \\
\textbf{SVM $-$ CRNN, full clip (primary)} & +1.3 & [$-$3.1, +5.7] & inconclusive \\
SVM (full clip) $-$ CRNN (1.5 s) & +3.0 & [$-$0.6, +6.5] & inconclusive \\
SVM $-$ CRNN, both 1.5 s & $-$0.7 & [$-$5.1, +3.7] & inconclusive \\
RF $-$ SVM & $-$1.3 & [$-$4.2, +1.6] & inconclusive \\
SVM + SMOTE & $-$0.2 & [$-$2.0, +1.5] & comparable \\
CRNN + SpecAugment & $-$2.9 & [$-$7.1, +1.3] & inconclusive \\
\bottomrule
\end{tabular}
\end{table}

\textbf{No architecture effect is resolved.} The standardized SVM, both RFs and both CRNNs reach 0.75--0.78 macro recall under grouped CV (Table~\ref{tab:main}). The SVM and the full-window CRNN differ by $+1.3$ points (95\% CI $-3.1$ to $+5.7$; Table~\ref{tab:compare}), by $-0.7$ with both on a 1.5\,s window, and RF and SVM by $-1.3$. These are inconclusive, not equal: differences of about five points either way cannot be excluded.

\textbf{Preprocessing and evaluation move results more than architecture does.} Clip-level CV inflates macro recall by 7.5 (SVM), 9.8 (RF) and 8.0 (CRNN) points. Dropping standardization costs the RBF SVM 26.9 points on identical features and folds, several times any architecture difference the data allow. Standardization is routine for RBF kernels, so this is no discovery about SVMs: one omitted step outweighs the choice of model family. RF class weights change nothing (0.767 vs.\ 0.766). With augmentation confined to training folds, SMOTE and waveform augmentation are comparable to none for the SVM, and SpecAugment is inconclusive.

\textbf{Recording source predicts class.} Native sampling rates are 4\,kHz (504 clips) or 44.1\,kHz (707), and rate and class are strongly associated (Cram\'er's $V=0.65$; Table~\ref{tab:rate}). A classifier given only the rate reaches 0.38 macro recall, near its ceiling of 0.40, since it can output only two classes. The 1.8\,kHz low-pass leaves the SVM comparable and the CRNN inconclusive ($-2.0$), so the SVM does not rely on the high band itself, and the CRNN at most modestly. Yet in an exploratory analysis that was not pre-registered (Table~\ref{tab:rate}), clips at a class's minority rate are mostly assigned to classes typical of that rate: the SVM labels 84\% of 4\,kHz pneumonia and 95\% of 4\,kHz Bronchial clips as asthma, and 75\% of 44.1\,kHz asthma clips as pneumonia or healthy. This holds for the SVM and the CRNN alike and after low-pass filtering, while COPD, the only class well represented at both rates, is recalled similarly at both. Minority cells hold only 4--7 patients, so the magnitudes are imprecise, but the direction is what a source shortcut predicts: the models learned recording provenance as well as disease.

\begin{table}[t]
\caption{Exploratory (post hoc): clips (patients) by class and native rate, and grouped-CV recall.}
\label{tab:rate}
\centering\small
\begin{tabular}{lcccccc}
\toprule
 & \multicolumn{2}{c}{Clips (patients)} & \multicolumn{2}{c}{SVM recall} & \multicolumn{2}{c}{CRNN recall} \\
\cmidrule(lr){2-3}\cmidrule(lr){4-5}\cmidrule(lr){6-7}
Class & 4\,kHz & 44.1\,kHz & 4\,kHz & 44.1\,kHz & 4\,kHz & 44.1\,kHz \\
\midrule
asthma & 262 (53) & 26 (6) & 0.92 & 0.20 & 0.91 & 0.14 \\
Bronchial & 15 (4) & 89 (19) & 0.00 & 0.83 & 0.16 & 0.75 \\
COPD & 188 (28) & 213 (31) & 0.83 & 0.89 & 0.83 & 0.82 \\
healthy & 0 & 133 (25) & n/a & 0.77 & n/a & 0.76 \\
pneumonia & 39 (7) & 246 (40) & 0.11 & 0.79 & 0.22 & 0.80 \\
\bottomrule
\end{tabular}
\end{table}

\section{What This Means for Deployment}

Neither accuracy nor latency separates the models. On a CPU-only laptop (Intel Core Ultra 7 155H), median single-clip time from file to prediction is 61\,ms for the SVM, 74\,ms for the RF and 51\,ms for the 1.5\,s CRNN the app ships (200 clips). What differs is footprint: the classical serving stack occupies 191\,MB on disk and PyTorch adds 449\,MB, which matters for slow downloads.
The prototype is packaged as an offline Tauri application around a local FastAPI inference service, so recordings never leave the device. Voice and breathing patterns are biometric data, and keeping them local removes the interception surface rather than mitigating it. The interface (Appendix~B) reports the full probability distribution and a standing disclaimer that the output is decision support requiring clinical confirmation. It ships the CRNN, though nothing here favors it over the classical models. More importantly, no model trained on ADD~v2 should be used on patients: a clinic recording everyone on one device removes their cue.

\section{Limitations}

\textbf{Recording provenance.} Class is entangled with recording source in ADD~v2: the source documents no device or site, and its filename codes never recur across classes. Our model comparison is therefore also made on a confounded benchmark; separating the two needs every class recorded on the same devices and sites. \textbf{Exploratory analysis.} The within-class analysis was not pre-registered and rests on 4--7 patients per minority cell, who may also differ clinically. \textbf{Identity and labels.} Filename patient numbers are imperfect, as the duplicates show; no diagnostic criteria are documented, so we keep the source label ``Bronchial'', and the dataset's provenance has been questioned~\cite{dolu}. \textbf{Untuned models.} Hyperparameters were fixed in advance; 18 of 125 CRNN runs hit the 50-epoch cap, though late-peaking runs scored no higher. \textbf{Population.} The source carries no geographic or demographic labels, so we do not know how any of these models perform on African patients, where biomass-smoke exposure, comorbidities such as HIV and tuberculosis, and recording practice may shift the acoustic distribution. \textbf{No prospective validation.} Nothing here has been tested with real patients, clinicians, or against gold-standard diagnosis. \textbf{Offline drift.} The application has no update channel, so a new model requires a reinstall.

\section{Conclusion}

On ADD~v2, under pre-registered patient-grouped evaluation, a standardized SVM, a random forest and a 1.13M-parameter CRNN cannot be told apart (0.75--0.78 macro recall), while a missing standardization step removes 27 points: feature scaling beats architecture choice. The more important finding lies in the data: recording source predicts class, and the models follow it. For a clinic without a GPU the classical models lose nothing measurable, but this corpus cannot tell us how any model would perform on patients. We release our harness so others can run the same two checks (a rate-by-class table and within-class recall by rate) before trusting a respiratory-sound benchmark.

\appendix

\section{Experimental setup}

All values are read from \texttt{experiments/config.py} and the results and environment records committed with them. Classical models ran on a Windows CPU machine and CRNNs on a Kaggle NVIDIA T4, both with librosa~0.11.0 and soxr~1.0.0.

\textbf{Deviations from the pre-registered plan.} (i)~Two code fixes followed the freeze without changing any computation: explicit UTF-8 output (commit \texttt{33715fe}, before any result) and re-rooting manifest paths across operating systems (\texttt{ad8c7ff}, after the classical results and before the CRNN runs).  (ii)~The latency script (\texttt{measure\_latency.py}, commit \texttt{eadf374}) was written after the freeze to implement the measurement the plan specifies.  (iii)~The within-class analysis by sampling rate (Table~3) is post hoc. It pools the committed grouped-protocol predictions over the five repeats and computes recall within each class and native sampling rate recorded in the manifest.

\begin{table}[H]
\caption{Configuration behind every number in the paper.}
\centering\footnotesize
\begin{tabular}{ll}
\toprule
\multicolumn{2}{l}{\emph{Data and protocol}}\\
Corpus & ADD~v2, 1,211 clips, 201 patient groups, 5 classes \\
Grouping & source label + filename patient number; groups joined by exact duplicates merged \\
Primary protocol & StratifiedGroupKFold, 5 folds $\times$ 5 seeds (25 runs), identical for all models \\
Secondary protocol & StratifiedKFold, 5 folds $\times$ 5 seeds, to measure leakage \\
Statistics & paired corrected resampled $t$-test~\cite{nadeau}; equivalence margin $\pm3$ points \\
Pre-registration & plan, manifest and folds committed before any run (commit \texttt{69ab10f}) \\
\midrule
\multicolumn{2}{l}{\emph{Features}}\\
Sample rate & 16\,kHz (librosa default resampler) \\
Classical & 30-D: mean and SD of 13 MFCCs, ZCR, RMS over the whole clip \\
CRNN input & log-Mel, $n_\text{fft}$ 2048, hop 512, 128 bins; 1.5\,s ($128\times47$) or 6.0\,s ($128\times188$) \\
Low-pass control & 10th-order zero-phase Butterworth at 1.8\,kHz, applied to all files \\
\midrule
\multicolumn{2}{l}{\emph{Models}}\\
SVM & RBF, $C=1$, $\gamma=$\,scale, balanced weights, Platt probabilities \\
RF & 100 trees, with or without balanced weights \\
CRNN & 1,134,789 parameters; Adam $10^{-3}$, batch 32, $\le$50 epochs, patience 10 \\
CRNN validation & 1 of 7 stratified patient-grouped folds of the training data \\
SMOTE & minority class only, fitted on the training fold \\
Waveform aug. & one copy per training clip: noise ($p$=0.7), pitch shift ($p$=0.5), time stretch ($p$=0.5) \\
SpecAugment & one frequency mask ($<$10 bins), one time mask ($<$15 frames), no time warp \\
\midrule
\multicolumn{2}{l}{\emph{Deployment benchmark (\texttt{measure\_latency.py}; CPU only)}}\\
Machine & Intel Core Ultra 7 155H, 33.8\,GB RAM, Windows; PyTorch 2.10 CPU build \\
Protocol & single-clip calls, 200 clips after 5 warm-up calls; median (p95) \\
End to end & SVM 61 (102)\,ms; RF 74 (126)\,ms; CRNN, 1.5\,s window, 51 (92)\,ms \\
Load + resample & 27 (45)\,ms, shared by all pipelines \\
Features & classical 32 (62)\,ms; CRNN log-Mel 13 (29)\,ms \\
Model & SVM 1.7 (4.1)\,ms; RF 13.7 (28.8)\,ms; CRNN 9.8 (29.4)\,ms \\
Footprint on disk & classical stack 191\,MB; with PyTorch 641\,MB (PyTorch 449\,MB) \\
CRNN checkpoint & 4.5\,MB \\
Updates & no update channel; a new model requires reinstall \\
\bottomrule
\end{tabular}
\end{table}

\section{The deployed interface}

\begin{figure}[H]
\centering
\includegraphics[width=0.9\linewidth]{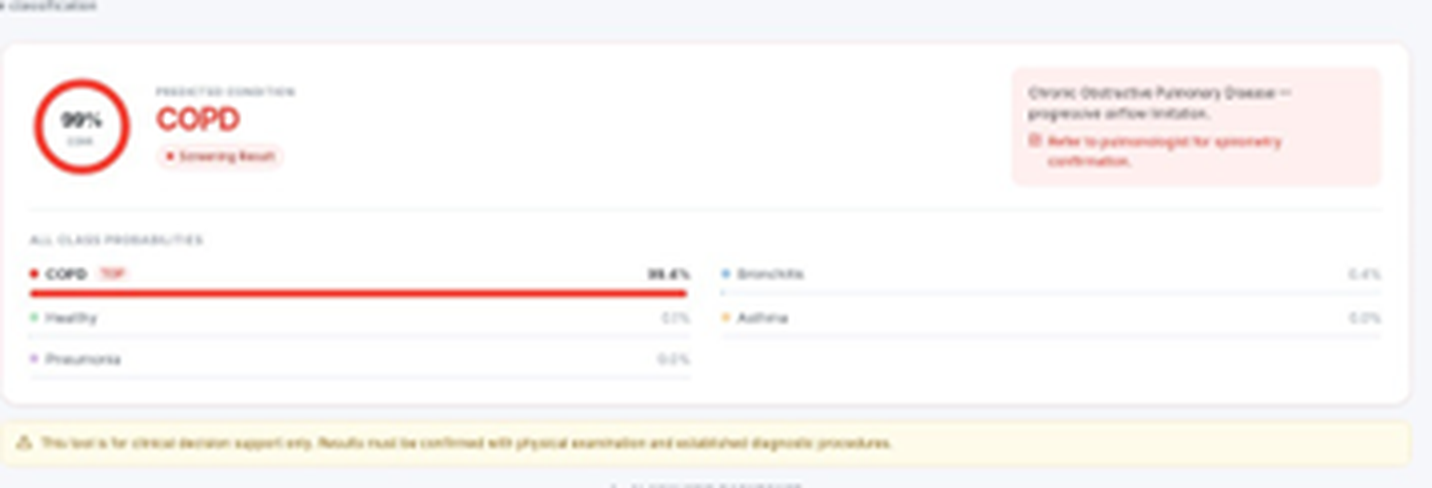}
\caption{The screening view running offline, showing the predicted class, confidence, per-class probabilities, referral note and the standing clinical disclaimer.}
\end{figure}

\begin{figure}[H]
\centering
\includegraphics[width=0.9\linewidth]{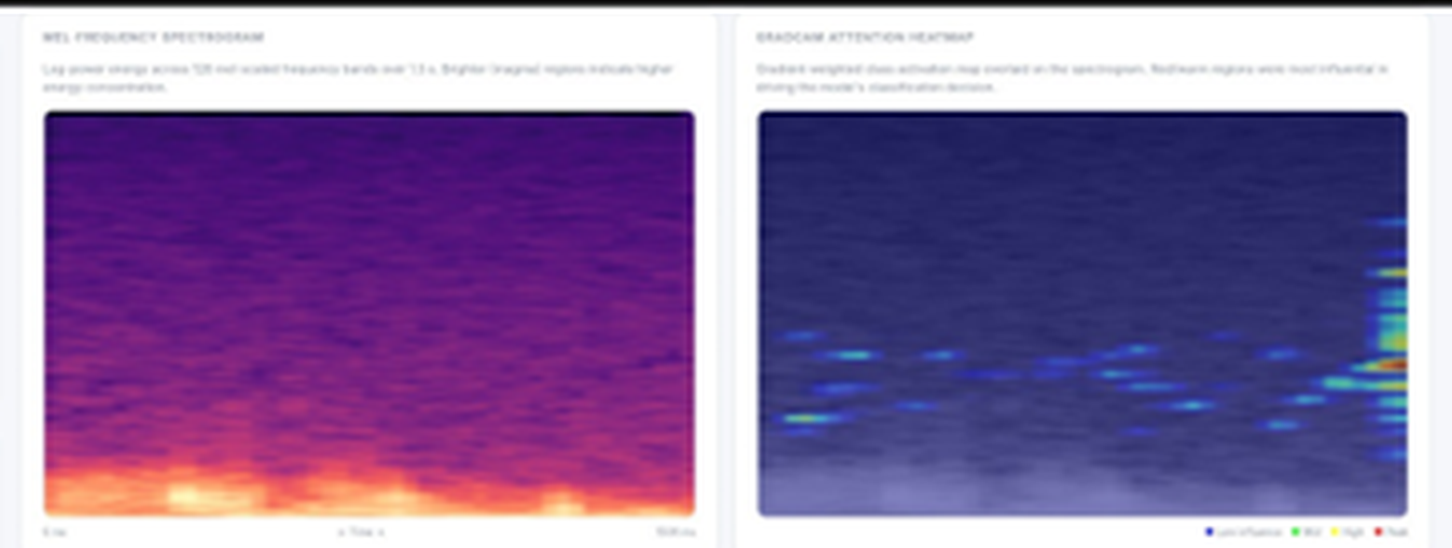}
\caption{The analysis view, showing the log-Mel spectrogram passed to the network (left) and a Grad-CAM attribution map over the same time-frequency axes (right). We have not evaluated whether these attributions are faithful, and the interface does not present them as an explanation.}
\end{figure}

\clearpage
\section{Complete results}

\begin{table}[H]
\caption{Every pre-registered configuration under both protocols: macro recall (mean $\pm$ SD over 25 runs) and mean per-class recall.}
\centering\scriptsize\setlength{\tabcolsep}{4pt}
\begin{tabular}{lllcccccc}
\toprule
ID & Configuration & Protocol & Macro recall & Asthma & Bronchial & COPD & Healthy & Pneumonia \\
\midrule
C1 & SVM, unscaled & grouped & 0.510 $\pm$ 0.053 & 0.72 & 0.48 & 0.60 & 0.21 & 0.54 \\
C1 & SVM, unscaled & clip & 0.542 $\pm$ 0.023 & 0.74 & 0.61 & 0.60 & 0.22 & 0.54 \\
C2 & SVM, standardized & grouped & 0.779 $\pm$ 0.034 & 0.85 & 0.71 & 0.86 & 0.77 & 0.70 \\
C2 & SVM, standardized & clip & 0.854 $\pm$ 0.025 & 0.86 & 0.81 & 0.91 & 0.91 & 0.78 \\
C3 & SVM, standardized, 26 features & grouped & 0.771 $\pm$ 0.040 & 0.84 & 0.70 & 0.85 & 0.78 & 0.68 \\
C3 & SVM, standardized, 26 features & clip & 0.845 $\pm$ 0.031 & 0.85 & 0.79 & 0.91 & 0.92 & 0.75 \\
C8 & SVM, standardized, first 1.5 s & grouped & 0.742 $\pm$ 0.051 & 0.83 & 0.65 & 0.85 & 0.73 & 0.65 \\
C8 & SVM, standardized, first 1.5 s & clip & 0.800 $\pm$ 0.027 & 0.84 & 0.75 & 0.88 & 0.81 & 0.72 \\
C4 & RF, unweighted & grouped & 0.767 $\pm$ 0.048 & 0.85 & 0.69 & 0.89 & 0.67 & 0.73 \\
C4 & RF, unweighted & clip & 0.870 $\pm$ 0.024 & 0.88 & 0.82 & 0.95 & 0.88 & 0.83 \\
C5 & RF, balanced weights & grouped & 0.766 $\pm$ 0.047 & 0.84 & 0.68 & 0.91 & 0.67 & 0.73 \\
C5 & RF, balanced weights & clip & 0.864 $\pm$ 0.028 & 0.88 & 0.79 & 0.95 & 0.85 & 0.84 \\
C6 & CRNN, 1.5 s window & grouped & 0.749 $\pm$ 0.042 & 0.81 & 0.71 & 0.77 & 0.71 & 0.74 \\
C6 & CRNN, 1.5 s window & clip & 0.829 $\pm$ 0.029 & 0.84 & 0.81 & 0.87 & 0.83 & 0.80 \\
C7 & CRNN, 6.0 s window & grouped & 0.766 $\pm$ 0.045 & 0.84 & 0.67 & 0.83 & 0.77 & 0.72 \\
A1 & SVM, + SMOTE & grouped & 0.777 $\pm$ 0.039 & 0.86 & 0.69 & 0.86 & 0.77 & 0.70 \\
A1 & SVM, + SMOTE & clip & 0.853 $\pm$ 0.026 & 0.86 & 0.80 & 0.91 & 0.90 & 0.78 \\
A3 & SVM, + waveform aug. & grouped & 0.778 $\pm$ 0.044 & 0.83 & 0.71 & 0.88 & 0.74 & 0.72 \\
A3 & SVM, + waveform aug. & clip & 0.858 $\pm$ 0.026 & 0.85 & 0.81 & 0.94 & 0.90 & 0.79 \\
A5 & SVM, + SMOTE + waveform aug. & grouped & 0.774 $\pm$ 0.047 & 0.83 & 0.69 & 0.88 & 0.74 & 0.72 \\
A5 & SVM, + SMOTE + waveform aug. & clip & 0.859 $\pm$ 0.027 & 0.85 & 0.81 & 0.94 & 0.89 & 0.81 \\
A2 & RF, + SMOTE & grouped & 0.766 $\pm$ 0.047 & 0.84 & 0.71 & 0.90 & 0.66 & 0.73 \\
A2 & RF, + SMOTE & clip & 0.873 $\pm$ 0.025 & 0.87 & 0.85 & 0.95 & 0.85 & 0.85 \\
A4 & RF, + waveform aug. & grouped & 0.764 $\pm$ 0.049 & 0.83 & 0.68 & 0.90 & 0.68 & 0.72 \\
A4 & RF, + waveform aug. & clip & 0.862 $\pm$ 0.026 & 0.87 & 0.82 & 0.95 & 0.84 & 0.83 \\
A6 & RF, + SMOTE + waveform aug. & grouped & 0.763 $\pm$ 0.040 & 0.81 & 0.70 & 0.89 & 0.67 & 0.73 \\
A6 & RF, + SMOTE + waveform aug. & clip & 0.865 $\pm$ 0.028 & 0.85 & 0.85 & 0.95 & 0.83 & 0.84 \\
A7 & CRNN, 1.5 s + SpecAugment & grouped & 0.720 $\pm$ 0.058 & 0.85 & 0.70 & 0.72 & 0.64 & 0.70 \\
B0 & Probe, native rate only & grouped & 0.382 $\pm$ 0.013 & 0.91 & 0.00 & 0.00 & 1.00 & 0.00 \\
B0 & Probe, native rate only & clip & 0.382 $\pm$ 0.008 & 0.91 & 0.00 & 0.00 & 1.00 & 0.00 \\
B1 & Probe, energy $>$2 kHz only & grouped & 0.313 $\pm$ 0.031 & 0.88 & 0.14 & 0.15 & 0.22 & 0.17 \\
B1 & Probe, energy $>$2 kHz only & clip & 0.318 $\pm$ 0.031 & 0.88 & 0.15 & 0.16 & 0.22 & 0.18 \\
B2\_C2 & SVM, standardized, low-pass & grouped & 0.778 $\pm$ 0.032 & 0.86 & 0.71 & 0.86 & 0.76 & 0.70 \\
B2\_C2 & SVM, standardized, low-pass & clip & 0.854 $\pm$ 0.024 & 0.88 & 0.81 & 0.91 & 0.91 & 0.77 \\
B2\_C5 & RF, balanced, low-pass & grouped & 0.765 $\pm$ 0.046 & 0.84 & 0.66 & 0.90 & 0.68 & 0.74 \\
B2\_C5 & RF, balanced, low-pass & clip & 0.863 $\pm$ 0.026 & 0.89 & 0.79 & 0.94 & 0.84 & 0.85 \\
B2\_C6 & CRNN, 1.5 s, low-pass & grouped & 0.729 $\pm$ 0.047 & 0.81 & 0.66 & 0.79 & 0.70 & 0.69 \\
\bottomrule
\end{tabular}
\end{table}

\begin{table}[H]
\caption{All pre-registered paired comparisons (patient-grouped CV).}
\centering\small
\begin{tabular}{lccl}
\toprule
Comparison & $\Delta$ (points) & 95\% CI & Verdict \\
\midrule
Standardization (SVM) & +26.9 & [+22.1, +31.6] & outperforms \\
30 vs.\ 26 features (SVM) & +0.8 & [$-$1.4, +3.1] & inconclusive \\
\textbf{SVM $-$ CRNN, full clip (primary)} & +1.3 & [$-$3.1, +5.7] & inconclusive \\
SVM (full clip) $-$ CRNN (1.5 s) & +3.0 & [$-$0.6, +6.5] & inconclusive \\
SVM $-$ CRNN, both 1.5 s & $-$0.7 & [$-$5.1, +3.7] & inconclusive \\
RF $-$ CRNN, full clip & 0.0 & [$-$5.7, +5.7] & inconclusive \\
RF $-$ SVM & $-$1.3 & [$-$4.2, +1.6] & inconclusive \\
SVM + SMOTE & $-$0.2 & [$-$2.0, +1.5] & comparable \\
SVM + waveform aug. & $-$0.1 & [$-$2.6, +2.4] & comparable \\
SVM + SMOTE + aug. & $-$0.5 & [$-$3.8, +2.8] & inconclusive \\
RF + SMOTE & +0.1 & [$-$2.4, +2.5] & comparable \\
RF + waveform aug. & $-$0.2 & [$-$3.0, +2.7] & inconclusive \\
RF + SMOTE + aug. & $-$0.3 & [$-$3.3, +2.6] & inconclusive \\
CRNN + SpecAugment & $-$2.9 & [$-$7.1, +1.3] & inconclusive \\
1.8 kHz low-pass, SVM & $-$0.2 & [$-$1.5, +1.2] & comparable \\
1.8 kHz low-pass, RF & $-$0.1 & [$-$2.2, +1.9] & comparable \\
1.8 kHz low-pass, CRNN & $-$2.0 & [$-$5.1, +1.1] & inconclusive \\
\bottomrule
\end{tabular}
\end{table}


{\small
\begin{thebibliography}{99}
\bibitem{who_asthma} World Health Organization. Asthma fact sheet, 2024.
\bibitem{gbd2021} GBD 2021 Lower Respiratory Infections and Antimicrobial Resistance Collaborators. Global, regional, and national incidence and mortality burden of non-COVID-19 lower respiratory infections and aetiologies, 1990--2021. \emph{The Lancet Infectious Diseases}, 24(9):974--1002, 2024.
\bibitem{who_copd} World Health Organization. Chronic obstructive pulmonary disease (COPD) fact sheet, 2024.
\bibitem{who_workforce} A.~Ahmat, S.~C. Okoroafor, I.~Kazanga, et al. The health workforce status in the WHO African Region: findings of a cross-sectional study. \emph{BMJ Global Health}, 7(Suppl~1):e008317, 2022.
\bibitem{rocha} B.~M. Rocha, et al. A respiratory sound database for the development of automated classification. In \emph{Precision Medicine Powered by pHealth and Connected Health}, pp.~51--55. Springer, 2018.
\bibitem{basu} V.~Basu, et al. Respiratory diseases recognition through respiratory sound with the help of deep neural network. arXiv:2104.04336, 2021.
\bibitem{moummad} I.~Moummad and N.~Farrugia. Pretraining respiratory sound representations using metadata and contrastive learning. In \emph{Proc.\ IEEE WASPAA}, pp.~1--5, 2023.
\bibitem{tawfik} M.~Tawfik, I.~S. Fathi, S.~S. Nimbhore, I.~M. Alsmadi, and M.~S. Sawah. E-RespiNet: An LLM-ELECTRA driven triple-stream CNN with feature fusion for asthma classification. \emph{PLOS ONE}, 20(11):e0334528, 2025.
\bibitem{saeb} S.~Saeb, L.~Lonini, A.~Jayaraman, D.~C. Mohr, and K.~P. Kording. The need to approximate the use-case in clinical machine learning. \emph{GigaScience}, 6(5):1--9, 2017.
\bibitem{zech} J.~R. Zech, M.~A. Badgeley, M.~Liu, A.~B. Costa, J.~J. Titano, and E.~K. Oermann. Variable generalization performance of a deep learning model to detect pneumonia in chest radiographs: A cross-sectional study. \emph{PLOS Medicine}, 15(11):e1002683, 2018.
\bibitem{degrave} A.~J. DeGrave, J.~D. Janizek, and S.-I. Lee. AI for radiographic COVID-19 detection selects shortcuts over signal. \emph{Nature Machine Intelligence}, 3:610--619, 2021.
\bibitem{geirhos} R.~Geirhos, J.-H. Jacobsen, C.~Michaelis, R.~Zemel, W.~Brendel, M.~Bethge, and F.~A. Wichmann. Shortcut learning in deep neural networks. \emph{Nature Machine Intelligence}, 2:665--673, 2020.
\bibitem{musaed} M.~T. Musaed. Asthma Detection Dataset Version 2 (ICBHI 2017). Kaggle, 2022. \url{https://www.kaggle.com/datasets/mohammedtawfikmusaed/asthma-detection-dataset-version-2}, accessed September 2026.
\bibitem{dolu} K.~O. Dolu. Comment on Topaloglu et al.\ ``Machine learning-driven lung sound analysis: Novel methodology for asthma diagnosis''. \emph{Advances in Respiratory Medicine}, 94(2):15, 2026.
\bibitem{quinn} J.~A. Quinn, R.~Nakasi, P.~K.~B. Mugagga, et al. Deep convolutional neural networks for microscopy-based point of care diagnostics. In \emph{Proc.\ Machine Learning for Healthcare}, 2016.
\bibitem{groh} M.~Groh, C.~Harris, L.~Soenksen, et al. Evaluating deep neural networks trained on clinical images in dermatology with the Fitzpatrick 17k dataset. In \emph{CVPR Workshops}, 2021.
\bibitem{chawla} N.~V. Chawla, K.~W. Bowyer, L.~O. Hall, and W.~P. Kegelmeyer. SMOTE: Synthetic minority over-sampling technique. \emph{Journal of Artificial Intelligence Research}, 16:321--357, 2002.
\bibitem{park} D.~S. Park, W.~Chan, Y.~Zhang, et al. SpecAugment: a simple data augmentation method for automatic speech recognition. In \emph{Proc.\ Interspeech}, pp.~2613--2617, 2019.
\bibitem{nadeau} C.~Nadeau and Y.~Bengio. Inference for the generalization error. \emph{Machine Learning}, 52(3):239--281, 2003.
\end{thebibliography}
}
\end{document}